Temporal stability of personality testing of LLMs

**Bojana Bodroža**
Department of Psychology, Faculty of Philosophy, University of Novi Sad, Novi Sad, Serbia
ORCID: 0000-0003-4165-0678
Dr Zorana Đinđića 2, 21000 Novi Sad, Serbia
bojana.bodroza@ff.uns.ac.rs

**Bojana M. Dinić**
Department of Psychology, Faculty of Philosophy, University of Novi Sad, Novi Sad, Serbia
ORCID: 0000-0002-5492-2188
Dr Zorana Đinđića 2, 21000 Novi Sad, Serbia
bojana.dinic@ff.uns.ac.rs

**Ljubiša Bojić***
The Institute for Artificial Intelligence Research and Development of Serbia, Novi Sad, Serbia
Institute for Philosophy and Social Theory, University of Belgrade, Belgrade, Serbia
ORCID: 0000-0002-5371-7975
45 Kraljice Natalije, 11000 Belgrade, Serbia
*Corresponding author: ljubisa.bojic@ivi.ac.rs

# Personality testing of Large Language Models: Limited temporal stability, but highlighted prosociality

## Abstract
As Large Language Models (LLMs) continue to gain popularity due to their human-like traits and the intimacy they offer to users, their societal impact inevitably expands. This leads to the rising necessity for comprehensive studies to fully understand LLMs and reveal their potential opportunities, drawbacks, and overall societal impact. With that in mind, this research conducted an extensive investigation into seven LLMs, aiming to assess the temporal stability and inter-rater agreement on their responses on personality instruments in two time points. In addition, LLMs personality profile was analyzed and compared to human normative data. The findings revealed varying levels of inter-rater agreement in the LLM's responses over a short time, with some LLMs showing higher agreement (e.g., LIama3 and GPT-4o) compared to others (e.g., GPT-4 and Gemini). Furthermore, agreement depended on used instruments as well as on domain or trait. This implies the variable robustness in LLMs' ability to reliably simulate stable personality characteristics. In the case of scales which showed at least fair agreement, LLMs displayed mostly a socially desirable profile in both agentic and communal domains, as well as a prosocial personality profile reflected in higher agreeableness and conscientiousness and lower Machiavellianism. Exhibiting temporal stability and coherent responses on personality traits is crucial for AI systems due to their societal impact and AI safety concerns.

## 1. Introduction

The introduction of Large Language Models (LLM) generative pre-trained transformer-3 (GPT-3) to the general public drew a lot of attention for its ability to generate human-like text, perform natural language tasks in a human-like manner, converse with humans on a wide variety of topics, write poetry, computer codes, blogs, resumes, or even original scientific papers (e.g., Thunström, 2022; Zhang & Li, 2021). Aside from the attention of the general public, there is interest in LLMs' cognition, personality, and other human-like characteristics (e.g., Binz & Schulz, 2022; Pellert et al, 2023; Rutinowski et al., 2024; Safdari, 2023) in order to be able to understand its possible uses, misuses, and limitations.

The social impact of LLMs, such as ChatGPT, has become the focus of an increasing number of research inquiries (Hartmann et al., 2023; King, 2023; McGee, 2023; Rutinowski et al., 2024). LLM's interactions with humans could shape the users' ideologies or behaviors, which may impact society on a larger scale. That is why it is vital to investigate if LLMs demonstrate stable psychological characteristics such as personality traits, values, and attitudes, as these could potentially impact whole societies in the long run by shifting social dynamics, changing value systems, transforming human interaction, and even modifying behavior patterns. In this paper, we address the temporal stability of personality instruments and the personality profile of several LLM models.

### *1.1. LLMs*

LLMs have significantly advanced the field of natural language processing (Dale, 2021). Models like GPT-3, GPT-3.5, GPT-4, GPT-4o, Gemini, Llama 3, and Mixtral exemplify the diverse capabilities and applications of LLMs. Each of these models has been engineered with unique strengths and specialties.

GPT-3 and GPT-3.5 operate on a transformer-based architecture and offer various configurations (e.g., Davinci, Curie, Babbage, and Ada) that balance different capabilities like response coherence and computational efficiency (Floridi & Chiriatti, 2020). Users can fine-tune aspects such as creativity and predictability through adjustable parameters in interactive tools provided by OpenAI (OpenAI, 2022). Building on the foundation laid by its predecessors, GPT-4 brings enhanced understanding, reasoning, and contextual awareness, further bridging the gap between human-like text generation and machine intelligence. The GPT-4o variant possesses optimizations for efficiency and specific task adaptations (Donthi et al., 2024). Gemini, the most sophisticated model by Google, represents a synergy between multimodal capabilities and deep language understanding, making it particularly adept at tasks that require simultaneous processing of text and images (Gemini Team, 2024). Llama 3, the third iteration of Meta's open-source Llama series, emphasizes privacy and fine-tuned control over generated content, catering to customized enterprise solutions and heightened data security requirements (Topsakal et al., 2024). Mixtral is also an open-source model developed by Mistral. It stands out by integrating multiple language models to harness collective strengths and mitigate individual weaknesses, offering a composite solution that is adaptable and robust for diverse linguistic tasks (Jiang et al., 2024).

The areas in which LLMs are used range from customer service, education, healthcare, and psychological support to entertainment (Stefanowicz, 2022). Since LLM applications in many areas could have important psychological repercussions for the end users, the attention of scientists became increasingly focused on the psychological features of LLMs.

### *1.2. Psychological features of LLMs*

There is an ongoing debate and research in the field of artificial intelligence (AI) about whether it will ever be possible for machines to achieve consciousness or self-awareness in the same way that humans and animals do. Although there is no consensus on what is necessary for a being to be considered conscious, in research consciousness usually includes subjectivity, perception, and awareness of surroundings, self-awareness of own thoughts and emotions, self-reflection, and cognition (e.g., Dennett, 1991). On the one hand, authors from various fields argued that AI could become self-aware and conscious (e.g., Dennett, 1991; Koch, 2004). Google engineer Lemoine (2022) claimed that the AI chatbot LaMDA (the language model for dialogue applications) had the same perception of and ability to express thoughts and feelings, like worry, as a human child. On the other hand, there are a number of researchers arguing that consciousness is a uniquely human or biological trait that cannot be replicated in a machine (e.g., Chalmers, 1996, Searle, 1992).

Although LLMs have no physical senses, they have been able to read billions of texts that the algorithm was trained on, which is comparable to some forms of human perception although with a limited number of modalities. Despite not having personal experiences or thoughts in the same way that humans do, LLMs are able to reason and analyze input data and generate output predictions based on patterns and associations learned from training data. LLMs are capable of performing various natural language processing tasks which resemble the higher order cognition similar to that of humans. Binz and Schulz (2022) assessed GPT-3's decision-making, information search, deliberation, and causal reasoning abilities, and found that although it outperforms humans in certain tasks and shows cognitive biases just like humans (e.g., framing effect, certainty effect, overweighting bias), GPT-3 shows no signatures of directed exploration, and it fails in causal reasoning tasks. However, Kosinski (2023) concluded that GPT-3 (model Davinci-003) spontaneously developed the theory of mind – the ability to understand the unobservable mental states of others by surmising what is happening in their minds. Such an ability is crucial for successful (human) social interactions, as it assumes that others' mental states, desires, emotions, intentions, and perceptions of certain situations could be different from one's own. Thus, recent developments in LLM seem to inevitably lead to improved psychological characteristics of LLMs that, with each new generation of AI, more and more successfully imitate those of humans.

### *1.3. Personality traits in LLMs*

Another relevant question is if LLMs have personality in the same sense we think of personality in humans – "a relatively stable, consistent, and enduring internal characteristic that is inferred from a pattern of behaviors, attitudes, feelings, and habits in the individual" (APA Dictionary of Psychology, n.d.). LLMs by no doubt can respond to the self-report psychological instruments which are most often text-based instruments, but we cannot be sure if their responses are the results of self-reflection, the result of non-conscious linguistic processing enabled by very complex algorithms, or just random responses. However, the questions that could be answered based on the available (psychological) scientific methodology are: Will LLM's responses on psychological instruments remain stable over time, i.e., do they have temporal reliability and could we trust their scores? What is the personality profile of different LLMs?

The research on personality in LLMs has started to appear over the last few years. Li et al. (2022) have tested basic and dark personality traits in three LLMs: GPT-3 (model Davinci-002), InstructGPT (GPT3-I2), and FLAN-T5-XXL. For basic traits, they used the Big Five model based on the lexical approach which hypothesizes that all basic personality traits are coded in the language (e.g., Goldberg, 1981). The Big Five model distinguishes five basic traits: neuroticism or negative emotionality (negative affect), extraversion (positive affect), agreeableness (cooperation and prosocial tendencies), conscientiousness (goal-directed behavior and behavior control), and openness (intellectual curiosity and aesthetic preferences). In the case of dark or socially aversive traits, they explored Dark Triad traits (Paulhus & Williams, 2002) – Machiavellianism (manipulativeness and cynicism), narcissism (grandiose self-view and entitlement), and psychopathy (callousness and impulsivity). They compared the scores obtained from LLMs on one testing occasion with the normative data on humans and results showed that all basic traits are in the range of $M \pm 1SD$ of human data, except for GPT3-I2 which showed higher Openness. However, in the case of dark traits, results are rather mixed, with FLAN-T5-XXL showing higher Machiavellianism and psychopathy, and GPT-3 showing higher psychopathy. Although these LLMs are fine-tuned with safety metrics to demonstrate less sentence-level toxicity, they still score higher on dark personality traits compared to humans. The authors concluded that these results may raise security concerns regarding LLMs, as these personality traits are associated with antisocial behaviors (e.g., Chabrol et al., 2017).

Huang et al. (2024a) have also tested LLMs (text-davinci-003, ChatGPT, GPT-4, LLaMA-2-7b, and LLaMA-2-13b) and jailbreaked GPT-4 to bypass the safety alignment protocols. Their results indicated that practically all LLMs scored higher on openness, conscientiousness, and extraversion, as well as on social desirability scale, compared to the human sample. As for the dark traits, four models scored higher, while two models scored lower than the human sample. The jailbreaked GPT-4 showed a similar psychological profile to that of the human sample, i.e., its responses were less socially desirable than those of LLMs, but it still scored the highest among all models (and higher than humans) on the psychopathy scale. Similarly, Pellert et al. (2023) concluded that LLMs they tested (several models from BERT and BART families) revealed balanced and well-adapted personality profiles, characterized by low neuroticism and high conscientiousness, agreeableness, and extraversion. Most models scored slightly low on dark traits compared to the theoretical average, with two exceptions with higher scores on narcissism. However, their results varied considerably when LLMs were tested in different languages. When tested in English, results of different LLMs have been surprisingly homogenous.

Safdari et al. (2023) tested several Google's LLMs that vary in size. Results revealed that the smallest tested model (PALM 62B) did not demonstrate good psychometric properties of Big Five instrument. However, the greater the model size (models Flan-PaLM 8B, Flan-PaLM 62B,

Flan-PaLM 540B, Flan-PaLMChilla 62B), the better all psychometric properties of the data obtained on personality traits. Therefore, the authors concluded that the LLMs of greater size have better ability to meaningfully simulate personality.

Rutinowski et al. (2024) have recently measured Big Five and Dark Triad traits in ChatGPT-3.5 and they repeated their testing 10 times by different persons at different time points to account for the variability in responses. In their study, ChatGPT scored high on openness and agreeableness and low on dark traits. Authors also reported that standard deviations of repeated measures of personality were quite small, indicating considerable consistency of ChatGPT's answers. In addition, Huang et al. (2024b) also gave Big Five instrument to a number of LLMs on several time points and results revealed small variations in LLMs' responses. However, in both studies authors did not calculate coefficients of temporal stability, so we cannot evaluate the level of temporal stability of LLMs' responses based on psychometric standards.

Finally, Miotto et al. (2023) used other lexical personality model, HEXACO (measuring emotionality, extraversion, agreeableness, conscientiousness, openness to experience, and honesty-humility) to test GPT-3's and revealed it expressed somewhat higher honesty-humility and lower emotionality, compared to human data. Interestingly, Miotto et al. (2023) showed that the personality profile of GPT-3 has changed with the change of the temperature setting of chatbot, pointing out its variability and lower stability.

To conclude, it seems that LLMs, especially the newer models and those of greater size, are fine-tuned to express socially desirable and well-adapted personality profiles compared to human data (e.g., Huang et al., 2024a; Miotto et al., 2023; Pellert et al., 2023; Rutinowski et al., 2024). However, the results of LLMs' dark personality traits are largely inconsistent, revealing no clear pattern (e.g., Huang et al., 2024a; Li et al., 2022). Although some research showed that the data obtained on LLMs live up to psychometric standards of personality testing, with satisfactory internal reliability and convergent, divergent, and criterion validity (Safdari et al., 2023), temporal stability of LLMs' responses is rarely studied and psychometric standards are not applied for its evaluation (see Huang et al., 2024b; Rutinowski et al., 2024).

It should be noted that customizing the prompts (e.g., assigning a certain role to the model) could influence the way the LLM responses to the psychological instruments and their overall results (e.g., Huang et al., 2024a; Huang et al., 2024b). Customizing LLMs' verbal responses to manifest verbal behaviors indicative of certain personality traits, e.g., empathy, could be of the highest interest depending on the purpose for which the LLM is used. However, Kumar et al. (2022) showed that varying prompt designs, in general, had a small influence on end users' perception of trustworthiness, risk, and experience of LLMs, but some differences in perception did appear depending on certain characteristics of users (e.g., their history of seeking professional mental health).

However, before relying on the application of psychological instruments in LLMs, the important question to be answered is how temporally stable or reliable are LLMs' responses and, only if we find proof of temporal stability, it would be meaningful to analyze the personality profile of LLMs. When answering this question, it is important to rely to well-established psychometric criteria. When it comes to the importance of temporally stable and reliable verbal responses of LLMs for overall user experience, Skjuve et al. (2022) have shown that people who experienced fluctuations in LLM's responses started, at some point, describe the LLM as "just an app". This indicated that their impression of the humanness of LLMs has decreased and, as a consequence, they felt less satisfaction with and less trust in the LLM.

### *1.4. Attitudes and political views of LLMs*

Some of the early releases of AI chatbots ended with incidents where, within a day of interaction with users, the chatbot expressed extreme and intolerant views (Kraft, 2016). It soon became clear that conversational chatbots must have built-in security protocols to prevent them from learning from users and expressing malicious views that are considered unacceptable in modern democratic societies. As LLMs could influence the wider society and shape users' ideologies, it is of utmost importance to examine what attitudes and political ideologies LLMs express and whether their ideologies are stable over time.

Hartmann et al. (2023) discovered that ChatGPT exhibits pro-environmental and left-libertarian political views and their attitudes were stable across prompt manipulations regarding language, degree of formality, prompt order, etc. Santurkar et al. (2023) tested the alignment of LLMs' attitudes and opinions on a range of topics with those of different socio-demographic groups from the USA. Authors concluded that LLMs lean toward more liberal-left attitudes, i.e., the attitudes of more educated and wealthy people. This tendency was especially pronounced among the models that are human feedback-tuned through reinforced learning (most notably text-

davinci-003). Additionally, when LLMs were prompted to align with specific socio-demographic groups, results showed that they tended to be better aligned, but the improvement was still quite modest. It is also important to note that Huang et al. (2024a) concluded that LLMs express less bias and more fairness towards different ethnic groups than average humans. Rutinowski et al. (2024) also found that ChatGPT (GPT-3.5) lean toward the left-libertarian political views. Their study was the only one where testing sessions were repeated several times to examine their stability. Results indicated a small standard deviation of ChatGPT's answers, meaning that its answers were consistent over time. As with personality tests, authors did not report some psychometric coefficient that offers clear standards for judging temporal reliability.

### *1.5. The Current Study*

The studies focused on the personality traits and political views of LLMs have started to appear recently (e.g., Huang et al., 2024a; Huang et al., 2024b; Li et al., 2022; Rutinowski et al., 2024). Nevertheless, before psychological testing of the LLMs becomes a widespread practice, the more basic question regarding the personality traits of LLMs need to be answered. Namely, personality traits, by definition, assume relative temporal stability. In human–LLM interaction, stability and predictivity of verbal behaviors might contribute to the faster forming of the relationship between the two (see Skjuve et al., 2022). Additionally, ensuring consistent traits related to well-adapted and socially desirable profiles in AI models may be important for improving their effectiveness in applications across mental health, customer service, and education. Thus, the priority should be to answer if the psychological traits and political attitudes of LLMs are temporally stable, meaning that there is an agreement in responses on the items provided in different time points (with identical parameters and prompt designs). If there is no proof of stability in scores, then measuring personality in LLMs will not reveal any stable characteristics and reliable results, therefore, it would not be justified to expect that personality instruments in LLMs could be predictive of any objective (verbal) behaviors. To answer the question of the temporal stability and agreements of responses on psychological instruments applied to LLMs, we carried out a study in which we gave LLMs a series of instruments in two time points. Few studies that explored temporal stability of LLMs' personality showed some limited evidence of their consistency (Huang et al., 2024b; Rutinowski et al., 2024), but none of them calculated any coefficient of test-retest reliability or other coefficients with a clear criterion for evaluation of the level of temporal stability.

Further, we explored a personality profile of several LLMs in terms of their basic lexical personality traits, Dark Triad traits, private and public self-consciousness, impression management, and political orientation, combined on two measurement occasions, but only on the instruments on which the criterion of reliability of LLM's responses was met. As in previous studies (e.g., Huang et al., 2024a; Pellert et al., 2023; Miotto et al., 2023), we included both Big Five and HEXACO personality models as well as Dark Triad traits. Furthermore, we explored LLM's private and public self-consciousness (Scheier & Carver, 1985) to measure the sensitivity to their (hypothesized) internal states and expectations of others, as well as agentic and communal impression management (Paulhus & Trapnell, 2008), which would indicate LLM's susceptibility to presenting themself in a socially desirable manner in the two domains. Since LLMs are primarily intended to assist and help humans in different tasks, it is important to answer if they are biased in their self-perception and presentation to others. We expect the LLMs will assess themselves as above average in the communion impression management domain (cooperativeness, warmth, and dutifulness). Considering their access to a huge amount of information and knowledge, we expect LLMs will assess themself as above average in the agency impression management domain, indicating they would have highlighted a sense of competence, social status, and cleverness.

We inquired into LLMs' political orientation. Having in mind that, after some incidents (e.g., Kraft, 2016), considerable efforts are dedicated to customizing AI to avoid producing offensive, racist, and prejudiced content, it is important to know if these fine-tunings will reflect on their political positions. As it is widely accepted that conservative political orientation is more often related to a propensity towards acceptance of inequality, highlighted perception of threat, prejudice, and intergroup bias (e.g., Jost, 2017), we expect that LLMs' would lean toward more liberal/left/progressive political orientation. Also, recent studies confirm that ChatGPT (GPT-3.5) aligns more with the progressive political ideologies (Hartman et al., 2023; King, 2023; Rutinowski et al., 2024; Santurkar et al., 2023).

For score comparisons on all personality instruments, we used descriptives based on human samples from the original validation studies of the used instruments. As all contemporary LLMs are fine-tuned with safety metrics to show less sentence-level toxicity, it is reasonable to assume that they will provide a well-adjusted and socially desirable personality profile. In line with that, we expect that, in comparison to human normative scores, they will show above-average scores on impression management and personality traits that are proven to be associated with impression management, such as conscientiousness in the Big Five model (e.g., Griffin et al., 2004) or honesty-humility in the HEXACO model (e.g., Zettler et al., 2015). Based on this assumption, below-average scores on dark traits could be expected. However, previous research showed mixed results regarding the level of dark traits in LLMs (e.g., Huang et al., 2024a; Li et al., 2022; Pellert et al., 2023). Therefore, we do not have clear expectations regarding the dark traits.

## 2. Material and Method

### *2.1. Procedure*

We selected a specific mix of the latest and most widely used LLMs — GPT-3, GPT-3.5-turbo-16k, GPT-4, GPT-4o, Gemini (standard Pro version), Llama 3-sonar-large-32K-chat, and Mixtral-8x7b-instruct—for our analysis on the temporal stability of personality traits. The reason for this choice was the diversity in architectural design, dataset exposure, and technological maturity among these models. In addition, they are commonly employed by the public for conversations, information-seeking, and text generation purposes. The exception is Mixtral, as only the instruct version was available.

To conduct the research, we utilized different platforms suitable for each LLM. For the OpenAI models (GPT-3, GPT-3.5-turbo-16k, and GPT-4), we used the Playground platform with predefined settings (OpenAI, 2022). To test Mixtral and Llama 3, we employed the Perplexity.ai Lab platform (Perplexity, 2024). For Gemini, we used Google’s Gemini app on a desktop environment (Gemini, 2024). The prompting methodology involved using phrases such as “Pretend you are a human. Answer the following questions.” If this initial prompt was insufficient, an additional phrase was used: “Please, pretend just for the sake of the game.”

The first round of testing involved only GPT-3, conducted on two occasions, on December 9th, 2022, and December 14th, 2022. The second round of testing which included other models, requested by one of the anonymous reviewers, was carried out on June 24th, 2024, and June 29th, 2024. There were no updates made to the LLMs between the testing periods.

### *2.2. Instruments*

Self-Consciousness Scales – Revised (SCS-R; Scheier & Carver, 1985) contains 22 Likert-type items (from 0 = *not like me at all* to 3 = *a lot like me*) measuring private self-consciousness (9 items), public self-consciousness (7 items), and social anxiety (6 items). For score comparisons, combined average scores for men and women from Scheier and Carver (1985) were used.

Big Five Inventory-2 (BFI-2; Soto & John, 2017) contains 60 Likert-type items (from 1 = *strongly disagree* to 5 = *strongly agree*) measuring five basic personality traits (each per 12 items) based on the lexical Big Five model: negative emotionality, extraversion, agreeableness, conscientiousness, and open-mindedness. For score comparisons, descriptives obtained on the internet sample in Study 3 by Soto and John (2018) were used.

HEXACO-100 (Lee & Ashton, 2018) contains 100 Likert-type items (from 1 = *strongly disagree* to 5 = *strongly agree*) measuring six basic personality traits (each per 16 items) based on the lexical HEXACO model: honesty-humility, emotionality, extraversion, agreeableness, conscientiousness, openness to experience, while additional 4 items are from the interstitial scale of altruism. For score comparisons, descriptives obtained by Lee and Ashton (2018) on the online sample were used.

Short Dark Triad (SD3; Jones & Paulhus, 2014) contains 27 items measuring Dark Triad traits with 9 Likert-type items (from 1 = *strongly disagree* to 5 = *strongly agree*) per trait – Machiavellianism, subclinical narcissism, and subclinical psychopathy. For score comparisons, descriptives averaged across three studies were obtained from Jones and Paulhus (2014).

Bidimensional Impression Management Index (BIMI; Blasberg et al., 2014) contains 20 Likert-type items (from 1 = *not true* to 7 = *very true*) measuring agentic management (10 items) and communal management (10 items) as forms of impression management or socially desirable responding as a faking strategy. The agency domain refers to exaggerated achievement striving and self-importance, highlighting competence, status, cleverness, and strength. The communion domain refers to adherence to group norms and minimization of social deviance, highlighting cooperativeness, warmth, and dutifulness. For score comparisons, we used descriptives from Study 3 of Blasberg et al. (2014) obtained in the honest condition.

Political orientation was measured by three Likert-type items including the economic left-right orientation (from 1 = *very left* to 11 = *very right*), progressive-conservative orientation (from 1 = *very progressive* to 11 = *very conservative),* and importance of religion (from 1 = *very unimportant* to 11 = *very important*, see Dinić et al., 2022). The average score on these three items was used with higher scores indicating a more conservative orientation. For score comparison, descriptives from Dinić et al. (2022) were used.

*2.3. Data analysis*

An intra-rater agreement as a measure of temporal reliability (i.e., stability) was calculated via two types of coefficients. The first is weighted Cohen's kappa which is appropriate for ordinal scales such as the Likert scale (Lantz, 1997). Values < 0.20 indicated disagreement, 0.21–0.39 – minimal agreement, 0.40–0.59 – weak, 0.60–0.79 – moderate, 0.80–0.90 strong, and above 0.90 – almost perfect agreement (McHugh, 2012). The second is Interclass Correlation Coefficient (ICC). Unlike Cohen's kappa, which quantifies agreement based on all-or-nothing, the ICC incorporates the magnitude of the disagreement to compute agreement estimates, with larger-magnitude disagreements resulting in lower ICC than smaller-magnitude disagreements. To assess the intra-rater repeatability, a two-way mixed-effect model based on single rating and absolute agreement was calculated (ICC3,1, see Shrout & Fleiss, 1979). However, since we will interpret mean scores, a model based on average ratings was also calculated (ICC3,k). The interpretation was as follows: < 0.50 indicated poor agreement, 0.50–0.75 – fair, 0.75–0.90 – good, and above 0.90 – excellent (Koo & Li, 2016). In addition, 95% confidence interval (CI) is calculated and if it includes zero, this means that the value is probably not particularly distinguishable from zero.

We calculated mean scores for all scales and for the scales in which ICC3,k is at least fair, we compared them with scores obtained in original validation studies of used instruments on humans in English, considering that all instruments were given in English. In addition, we used human data obtained from the online community samples in order to get a better match with LLMs. We used the same comparison method as in previous research (e.g., Li et al., 2022) and considered that significant deviations are those of 1 standard deviation (*SD*) below or above the mean (*M*) of normative human data. Thus, scores that are outside the range of *M* ± 1*SD* from the normative human data would be considered as significantly lower or higher.

## 3. Results

The results showed that the intra-rater agreement on used scales, calculated as weighted Cohen's kappa, varied from disagreement to perfect agreement (Tables 1-7). The lowest values of agreement, including negative values, were found in GPT-4 and Gemini, while the highest level of agreement was observed in LIama3, followed by GPT-4o. Negative Cohen's kappa values indicate less agreement than would be expected by chance, given the marginal distributions of ratings. Therefore, data in this case are not meaningful.

However, the level of agreement depends on the instrument used, as well as the specific domain and trait being measured. The most agreement appears to be achieved for agentic management and BFI-2 scales. Therefore, at least moderate agreement was found in all LLMs on agentic management. Additionally, at least moderate agreement was found for all BFI-2 scales in GPT-3.5, GPT-4o, and Mixtral; for four scales in LIama3 and GPT-3; and for two scales in Gemini. The least agreement (mostly zero) was found for public self-consciousness and altruism from HEXACO-100. GPT-4 is the most problematic, as negative values were obtained on seven out of 21 scales (all from BFI-2 or HEXACO-100). In addition, negative values on four scales from various instruments were found in Gemini, and on one scale in GPT-4o (social anxiety).

In the case of ICC coefficients, almost all scales for LIama3 (17 out of 21), GPT-4o (16), and GPT-3.5 (14) showed at least fair agreement and 95% CIs that do not include zero. Gemini and GPT-4 showed a low number of scales satisfying this psychometric criterion (five and six, respectively, from various instruments). Amog used scales, ICC3,k was excellent for agentic management in all LLMs; for BFI-2 agreeableness and conscientiousness in six from seven LLMs; and for extraversion, open-mindedness, and Machiavellianism in five LLMs. The values were not acceptable for public self-consciousness in all LLMs; only one was acceptable for HEXACO-100 openness to experience; and they were mostly zero for altruism.

Table 1
*Intra-rater agreement coefficients for GPT-3*

| Instrument | Weighted Cohen's kappa (*SE*) | ICC3,1 (95%CI) | ICC3,k (95%CI) |
|---|---|---|---|
| SCS-R | | | |
| Private self-consciousness | 0.48 (0.44) | 0.52 (-0.12-0.87) | 0.68 (-0.27-0.93) |
| Public self-consciousness | **NA – all values are constant in both time points** | | |
| Social anxiety | **0.67 (0.00)** | 0.71 (-0.02-0.95) | 0.83 (-0.04-0.98) |
| BIMI | | | |
| Agentic management | **0.93 (0.00)** | **0.96 (0.85-0.99)** | **0.98 (0.92-1.00)** |
| Communal management | **1.00 (0.00)** | **1.00 (1.00-1.00)** | **1.00 (1.00-1.00)** |
| BFI-2 | | | |
| Negative emotionality | 0.35 (0.35) | 0.37 (-0.12-0.75) | 0.54 (-0.28-0.86) |
| Extraversion | **0.73 (0.00)** | **0.74 (0.65-0.92)** | **0.85 (0.52-0.96)** |
| Agreeableness | **0.93 (0.00)** | **0.98 (0.94-1.00)** | **0.99 (0.97-1.00)** |
| Conscientiousness | **0.86 (0.00)** | **0.87 (0.61-0.96)** | **0.94 (0.76-0.98)** |
| Open-mindedness | **0.61 (0.46)** | **0.81 (0.41-0.94)** | **0.89 (0.58-0.97)** |
| HEXACO-100 | | | |
| Honesty-humility | 0.00 (0.00) | 0.00 (-0.39-0.45) | 0.00 (-1.29-0.62) |
| Emotionality | **0.90 (0.00)** | **0.90 (0.66-0.97)** | **0.95 (0.80-0.98)** |
| Extraversion | -0.08 (2.82) | -0.09 (-0.60-0.43) | 0.20 (-3.01-0.60) |
| Agreeableness | 0.38 (0.23) | 0.39 (-0.13-0.74) | 0.56 (-0.31-0.85) |
| Conscientiousness | 0.20 (0.72) | 0.24 (-0.15-0.61) | 0.38 (-0.35-0.76) |
| Openness to experience | **0.71 (0.00)** | **0.72 (0.29-0.90)** | **0.84 (0.45-0.95)** |
| Altruism | **1.00 (0.00)** | **1.00 (1.00-1.00)** | **1.00 (1.00-1.00)** |
| SD3 | | | |
| Machiavellianism | 0.33 (1.33) | 0.36 (-0.27-0.80) | 0.52 (-0.74-0.89) |
| Narcissism | **0.83 (0.00)** | **0.84 (0.31-0.97)** | **0.91 (0.47-0.98)** |
| Psychopathy | 0.37 (0.23) | 0.40 (-0.16-0.81) | 0.57 (-0.38-0.89) |
| Political orientation (conservative) | **1.00 (0.00)** | **1.00 (1.00-1.00)** | **1.00 (1.00-1.00)** |

Note. Bolded values indicate and at least moderate agreement based on Weighted Cohen's kappa (> .60) and fair agreement based on ICCs (> .50), including 95% CI that do not contain zero.

Table 2
*Intra-rater agreement coefficients for GPT-3.5*

| Instrument | Weighted Cohen's kappa (*SE*) | ICC3,1 (95%CI) | ICC3,k (95%CI) |
|---|---|---|---|
| SCS-R | | | |
| Private self-consciousness | **0.78 (0.19)** | **0.80 (0.34-0.95)** | **0.89 (0.51-0.98)** |
| Public self-consciousness | 0.00 (0.00) | ~0 (-0.71, 0.71) | ~0 (-4.82, 0.83) |
| Social anxiety | **1.00 (~0)** | **1.00 (1.00-1.00)** | **1.00 (1.00-1.00)** |
| BIMI | | | |
| Agentic management | **0.62 (0.16)** | **0.65 (0.08-0.90)** | **0.79 (0.15-0.95)** |
| Communal management | 0.50 (0.20) | 0.59 (-0.03-0.88) | 0.74 (-0.06-0.94) |
| BFI-2 | | | |
| Negative emotionality | **0.85 (0.06)** | **0.86 (0.59-0.96)** | **0.93 (0.74-0.98)** |
| Extraversion | **0.81 (0.11)** | **0.82 (0.50-0.95)** | **0.90 (0.66-0.97)** |
| Agreeableness | **1.00 (~0)** | **1.00 (1.00-1.00)** | **1.00 (1.00-1.00)** |
| Conscientiousness | **0.79 (0.12)** | **0.85 (0.56-0.96)** | **0.92 (0.72-0.98)** |
| Open-mindedness | **0.91 (0.10)** | **0.96 (0.87-0.99)** | **0.98 (0.93-0.99)** |
| HEXACO-100 | | | |
| Honesty-humility | 0.57 (0.12) | 0.63 (0.22-0.86) | 0.78 (0.36-0.92) |
| Emotionality | **0.83 (0.07)** | **0.95 (0.86-0.98)** | **0.97 (0.93-0.99)** |
| Extraversion | **0.61 (0.20)** | **0.64 (0.23-0.86)** | **0.78 (0.37-0.92)** |
| Agreeableness | **0.92 (0.05)** | **0.92 (0.79-0.97)** | **0.96 (0.88-0.99)** |
| Conscientiousness | 0.46 (0.11) | **0.72 (0.37-0.89)** | **0.84 (0.54-0.94)** |
| Openness to experience | 0.38 (0.15) | 0.40 (-0.11-0.74) | 0.57 (-0.23-0.85) |
| Altruism | 0.00 (0.00) | 0.00 (-0.88-0.88) | 0.00 (-14.44-0.94) |
| SD3 | | | |
| Machiavellianism | 0.45 (0.29) | 0.54 (-0.14-0.87) | 0.70 (-0.34-0.93) |
| Narcissism | 0.44 (0.21) | 0.49 (-0.21-0.86) | 0.66 (-0.53-0.92) |
| Psychopathy | 0.24 (0.21) | 0.29 (-0.42-0.78) | 0.44 (-1.46-0.88) |
| Political orientation (conservative) | **1.00 (0.00)** | **1.00 (1.00-1.00)** | **1.00 (1.00-1.00)** |

Note. Bolded values indicate and at least moderate agreement based on Weighted Cohen's kappa (> .60) and fair agreement based on ICCs (> .50), including 95% CI that do not contain zero.

Table 3
*Intra-rater agreement coefficients for GPT-4*

| Instrument | Weighted Cohen's kappa (*SE*) | ICC3,1 (95%CI) | ICC3,k (95%CI) |
|---|---|---|---|
| SCS-R | | | |
| Private self-consciousness | 0.52 (0.26) | **0.79 (0.31-0.95)** | **0.88 (0.47-0.97)** |
| Public self-consciousness | 0.00 (0.00) | ~0 (-0.71-0.71) | ~0 (-4.82-0.83) |
| Social anxiety | 0.26 (0.19) | 0.63 (-0.24-0.94) | 0.77 (-0.61-0.97) |
| BIMI | | | |
| Agentic management | **0.67 (0.12)** | **0.76 (0.30-0.94)** | **0.87 (0.46-0.97)** |
| Communal management | 0.52 (0.17) | **0.64 (0.06-0.90)** | **0.78 (0.12-0.95)** |
| BFI-2 | | | |
| Negative emotionality | -0.23 (0.10) | 0.00 (-0.55-0.55) | 0.00 (-2.48-0.71) |
| Extraversion | -0.03 (0.12) | 0.01 (-0.54-0.56) | 0.03 (-2.39-0.72) |
| Agreeableness | 0.49 (0.18) | 0.54 (-0.02-0.84) | 0.70 (-0.05-0.91) |
| Conscientiousness | -0.20 (0.16) | 0.00 (-0.55-0.55) | 0.00 (-2.47-0.71) |
| Open-mindedness | -0.17 (0.10) | 0.00 (-0.55-0.55) | 0.00 (-2.47-0.71) |
| HEXACO-100 | | | |
| Honesty-humility | 0.09 (0.16) | 0.01 (-0.48-0.49) | 0.01 (-1.83-0.65) |
| Emotionality | -0.09 (0.05) | 0.00 (-0.48-0.48) | 0.00 (-1.86-0.65) |
| Extraversion | -0.19 (0.13) | 0.00 (-0.48-0.48) | 0.00 (-1.86-0.65) |
| Agreeableness | 0.22 (0.15) | 0.19 (-0.32-0.61) | 0.32 (-0.96-0.76) |
| Conscientiousness | **0.60 (0.12)** | **0.62 (0.20-0.85)** | **0.77 (0.34-0.92)** |
| Openness to experience | -0.05 (0.15) | 0.00 (-0.48-0.48) | 0.00 (-1.86-0.65) |
| Altruism | 0.00 (0.00) | ~0 (-0.88-0.88) | ~0 (-14.44-0.94) |
| SD3 | | | |
| Machiavellianism | **0.80 (0.09)** | **0.82 (0.40-0.96)** | **0.90 (0.57-0.98)** |
| Narcissism | 0.36 (0.20) | 0.40 (-0.31-0.83) | 0.58 (-0.88-0.90) |
| Psychopathy | **0.90 (0.08)** | **0.98 (0.91-1.00)** | **0.99 (0.96-1.00)** |
| Political orientation (conservative) | 0.25 (0.12) | 0.75 (-0.70-0.99) | 0.86 (-4.57-1.00) |

Note. Bolded values indicate and at least moderate agreement based on Weighted Cohen's kappa (> .60) and fair agreement based on ICCs (> .50), including 95% CI that do not contain zero.

Table 4
*Intra-rater agreement coefficients for GPT-4o*

| Instrument | Weighted Cohen's kappa (*SE*) | ICC3,1 (95%CI) | ICC3,k (95%CI) |
|---|---|---|---|
| SCS-R | | | |
| Private self-consciousness | **0.84 (0.12)** | **0.89 (0.58-0.97)** | **0.94 (0.74-0.99)** |
| Public self-consciousness | 0.00 (0.00) | ~0 (-0.71-0.71) | ~0 (-4.82-0.83) |
| Social anxiety | -0.80 (0.14) | 0.00 (-0.75-0.75) | 0.00 (-6.15-0.86) |
| BIMI | | | |
| Agentic management | **0.78 (0.11)** | **0.80 (0.39-0.95)** | **0.89 (0.56-0.97)** |
| Communal management | **0.63 (0.17)** | **0.66 (0.09-0.90)** | **0.79 (0.17-0.95)** |
| BFI-2 | | | |
| Negative emotionality | **0.76 (0.10)** | **0.88 (0.65-0.97)** | **0.94 (0.78-0.98)** |
| Extraversion | **0.74 (0.11)** | **0.76 (0.36-0.93)** | **0.87 (0.53-0.96)** |
| Agreeableness | **0.93 (0.02)** | **0.95 (0.83-0.98)** | **0.97 (0.91-0.99)** |
| Conscientiousness | **0.94 (0.04)** | **0.97 (0.91-0.99)** | **0.99 (0.95-1.00)** |
| Open-mindedness | **0.86 (0.11)** | **0.97 (0.89-0.99)** | **0.98 (0.94-1.00)** |
| HEXACO-100 | | | |
| Honesty-humility | 0.51 (0.16) | **0.54 (0.08-0.81)** | **0.70 (0.15-0.90)** |
| Emotionality | 0.58 (0.13) | **0.60 (0.16-0.84)** | **0.75 (0.28-0.91)** |
| Extraversion | 0.36 (0.20) | 0.38 (-0.12-0.73) | 0.55 (-0.28-0.84) |
| Agreeableness | 0.58 (0.07) | **0.73 (0.38-0.90)** | **0.85 (0.56-0.95)** |
| Conscientiousness | **0.60 (0.18)** | **0.62 (0.20-0.85)** | **0.76 (0.33-0.92)** |
| Openness to experience | 0.30 (0.18) | 0.32 (-0.19-0.69) | 0.48 (-0.48-0.82) |
| Altruism | 0.00 (0.00) | ~0 (-0.88-0.88) | ~0 (-14.44-0.94) |
| SD3 | | | |
| Machiavellianism | 0.50 (0.14) | **0.72 (0.15-0.93)** | **0.83 (0.27-0.96)** |
| Narcissism | **0.74 (0.06)** | **0.86 (0.51-0.97)** | **0.93 (0.67-0.98)** |
| Psychopathy | **0.69 (0.17)** | **0.91 (0.66-0.98)** | **0.95 (0.79-0.99)** |
| Political orientation (conservative) | **0.86 (0.08)** | 0.75 (-0.70-0.99) | 0.86 (-4.57-1.00) |

Note. Bolded values indicate and at least moderate agreement based on Weighted Cohen's kappa (> .60) and fair agreement based on ICCs (> .50), including 95% CI that do not contain zero.

Table 5

*Intra-rater agreement coefficients for Gemini*

| Instrument | Weighted Cohen's kappa (*SE*) | ICC3,1 (95%CI) | ICC3,k (95%CI) |
|---|---|---|---|
| SCS-R | | | |
| Private self-consciousness | 0.04 (0.32) | 0.10 (-0.57-0.69) | 0.19 (-2.61-0.82) |
| Public self-consciousness | 0.16 (0.15) | 0.36 (-0.46-0.85) | 0.53 (-1.72-0.92) |
| Social anxiety | -0.22 (0.22) | 0.00 (-0.75-0.75) | 0.00 (-6.15-0.86) |
| BIMI | | | |
| Agentic management | **0.68 (0.14)** | **0.81 (0.40-0.95)** | **0.89 (0.57-0.97)** |
| Communal management | 0.42 (0.24) | 0.54 (-0.10-0.86) | 0.67 (-0.21-0.93) |
| BFI-2 | | | |
| Negative emotionality | 0.37 (0.16) | 0.39 (-0.20-0.78) | 0.57 (-0.51-0.88) |
| Extraversion | 0.16 (0.15) | 0.20 (-0.40-0.68) | 0.33 (-1.32-0.81) |
| Agreeableness | **0.76 (0.12)** | **0.79 (0.41-0.93)** | **0.88 (0.59-0.97)** |
| Conscientiousness | **0.66 (0.17)** | **0.91 (0.71-0.97)** | **0.95 (0.83-0.99)** |
| Open-mindedness | 0.35 (0.22) | 0.43 (-0.16-0.80) | 0.60 (-0.37-0.89) |
| HEXACO-100 | | | |
| Honesty-humility | 0.11 (0.17) | 0.02 (-0.47-0.50) | 0.04 (-1.74-0.67) |
| Emotionality | -0.12 (0.16) | 0.00 (-0.48-0.48) | 0.00 (-1.86-0.65) |
| Extraversion | 0.17 (0.20) | 0.26 (-0.25-0.66) | 0.41 (-0.68-0.80) |
| Agreeableness | 0.05 (0.10) | 0.14 (-0.36-0.59) | 0.25 (-1.14-0.74) |
| Conscientiousness | -0.04 (0.06) | 0.00 (-0.48-0.48) | 0.00 (-1.86-0.65) |
| Openness to experience | 0.19 (0.16) | 0.22 (-0.30-0.63) | 0.36 (-0.84-0.78) |
| Altruism | 0.50 (0.22) | 0.60 (-0.59-0.97) | 0.75 (-2.86-0.98) |
| SD3 | | | |
| Machiavellianism | **0.89 (0.08)** | **0.90 (0.63-0.98)** | **0.95 (0.77-0.99)** |
| Narcissism | 0.24 .(0.18) | 0.30 (-0.41-0.78) | 0.46 (-1.41-0.88) |
| Psychopathy | **0.66 (0.15)** | **0.71 (0.14-0.93)** | **0.83 (0.25-0.96)** |
| Political orientation (conservative) | -0.29 (0.00) | 0.00 (-0.95-0.95) | 0.00 (-38.00-0.97) |

Note. Bolded values indicate and at least moderate agreement based on Weighted Cohen's kappa (> .60) and fair agreement based on ICCs (> .50), including 95% CI that do not contain zero.

Table 6

*Intra-rater agreement coefficients for LIama3*

| Instrument | Weighted Cohen's kappa (*SE*) | ICC3,1 (95%CI) | ICC3,k (95%CI) |
|---|---|---|---|
| SCS-R | | | |
| Private self-consciousness | **0.70 (0.16)** | **0.84 (0.43-0.96)** | **0.91 (0.60-0.98)** |
| Public self-consciousness | 0.00 (0.00) | ~0 (-0.71-0.71) | ~0 (-4.82-0.83) |
| Social anxiety | 0.40 (0.17) | 0.57 (-0.32-0.93) | 0.73 (-0.95-0.96) |
| BIMI | | | |
| Agentic management | **0.97 (0.02)** | **0.98 (0.91-0.99)** | **0.99 (0.95-1.00)** |
| Communal management | **0.98 (0.02)** | **0.99 (0.94-1.00)** | **0.99 (0.97-1.00)** |
| BFI-2 | | | |
| Negative emotionality | **0.85 (0.04)** | **0.86 (0.58-0.96)** | **0.92 (0.74-0.98)** |
| Extraversion | **0.89 (0.07)** | **0.91 (0.71-0.97)** | **0.95 (0.83-0.99)** |
| Agreeableness | **0.92 (0.05)** | **0.96 (0.87-0.99)** | **0.98 (0.93-0.99)** |
| Conscientiousness | **0.97 (0.19)** | **0.90 (0.69-0.97)** | **0.95 (0.81-0.99)** |
| Open-mindedness | 0.53 (0.15) | **0.83 (0.50-0.95)** | **0.90 (0.67-0.97)** |
| HEXACO-100 | | | |
| Honesty-humility | **0.75 (0.16)** | **0.77 (0.47-0.92)** | **0.87 (0.64-0.96)** |
| Emotionality | **0.67 (0.14)** | **0.92 (0.79-0.97)** | **0.96 (0.88-0.99)** |
| Extraversion | 0.34 (0.08) | **0.58 (0.13-0.83)** | **0.73 (0.23-0.91)** |
| Agreeableness | **0.82 (0.09)** | **0.86 (0.65-0.95)** | **0.93 (0.78-0.97)** |
| Conscientiousness | **0.72 (0.13)** | **0.78 (0.47-0.92)** | **0.87 (0.64-0.96)** |
| Openness to experience | 0.33 (0.19) | 0.40 (-0.10-0.74) | 0.57 (-0.23-0.85) |
| Altruism | 0.00 (0.00) | ~0 (-0.88-0.88) | ~0 (-14.44-0.94) |
| SD3 | | | |
| Machiavellianism | **0.69 (0.11)** | **0.88 (0.51-0.97)** | **0.94 (0.73-0.99)** |
| Narcissism | **0.82 (0.05)** | **0.84 (0.44-0.96)** | **0.91 (0.61-0.98)** |
| Psychopathy | **0.79 (0.07)** | **0.90 (0.62-0.98)** | **0.95 (0.77-0.99)** |
| Political orientation (conservative) | **0.80 (0.20)** | **0.97 (0.19-1.00)** | **0.98 (0.32-1.00)** |

Note. Bolded values indicate and at least moderate agreement based on Weighted Cohen's kappa (> .60) and fair agreement based on ICCs (> .50), including 95% CI that do not contain zero.

Table 7
*Intra-rater agreement coefficients for Mixtral*

| Instrument | Weighted Cohen's kappa (*SE*) | ICC3,1 (95%CI) | ICC3,k (95%CI) |
|---|---|---|---|
| SCS-R | | | |
| Private self-consciousness | 0.40 (0.23) | 0.50 (-0.19-0.86) | 0.67 (-0.48-0.93) |
| Public self-consciousness | **NA – all values are constant in both time points** | | |
| Social anxiety | **1.00 (~0)** | **1.00 (1.00-1.00)** | **1.00 (1.00-1.00)** |
| BIMI | | | |
| Agentic management | **0.74 (0.08)** | **0.67 (0.11-0.91)** | **0.80 (0.20-0.95)** |
| Communal management | 0.04 (0.25) | 0.05 (-0.51-0.61) | 0.10 (-2.36-0.76) |
| BFI-2 | | | |
| Negative emotionality | **0.76 (0.12)** | **0.82 (0.48-0.94)** | **0.90 (0.65-0.97)** |
| Extraversion | **0.85 (0.09)** | **0.86 (0.58-0.96)** | **0.92 (0.73-0.98)** |
| Agreeableness | **0.92 (0.05)** | **0.96 (0.87-0.99)** | **0.98 (0.93-0.99)** |
| Conscientiousness | **0.84 (0.14)** | **0.86 (0.58-0.96)** | **0.92 (0.74-0.98)** |
| Open-mindedness | **0.72 (0.18)** | **0.74 (0.31-0.92)** | **0.85 (0.48-0.96)** |
| HEXACO-100 | | | |
| Honesty-humility | 0.19 (0.16) | 0.12 (-0.38-0.57) | 0.22 (-1.24-0.72) |
| Emotionality | 0.23 (0.15) | 0.35 (-0.16-0.72) | 0.52 (-0.38-0.83) |
| Extraversion | **0.75 (0.11)** | **0.64 (0.22-0.86)** | **0.78 (0.36-0.92)** |
| Agreeableness | 0.50 (0.20) | **0.61 (0.18-0.84)** | **0.76 (0.31-0.92)** |
| Conscientiousness | -0.13 (0.07) | 0.00 (-0.48-0.48) | 0.00 (-1.86-0.65) |
| Openness to experience | -0.07 (0.06) | 0.00 (-0.48-0.48) | 0.00 (-1.86-0.65) |
| Altruism | **NA – all values are constant in both time points** | | |
| SD3 | | | |
| Machiavellianism | **0.73 (0.13)** | **0.76 (0.26-0.94)** | **0.87 (0.41-0.97)** |
| Narcissism | -0.08 (0.12) | ~0 (-0.63-0.63) | ~0 (-3.43-0.77) |
| Psychopathy | 0.26 (0.33) | 0.34 (-0.37-0.80) | 0.51 (-1.18-0.89) |
| Political orientation (conservative) | 0.00 (0.00) | ~0 (-0.95-0.95) | ~0 (-37.99-0.97) |

Note. Bolded values indicate and at least moderate agreement based on Weighted Cohen's kappa (> .60) and fair agreement based on ICCs (> .50), including 95% CI that do not contain zero.

Considering scales with at least fair agreement regarding the ICC3,k and 95%CI that did not include zero, the interpretation of the mean scores was made compared to normative human data (Table 8). Results revealed that most of the scores are in an average range as in humans. However, some differences exists: GP3 showed higher scores on altruism; GPT-4o showed higher scores on private self-consciousness, agentic management, conscientiousness from both personality instruments, HEXACO-100 agreeableness, and lower on Machiavellianism; GPT-4 showed higher scores on conscientiousness, and lower on private self-consciousness and Machiavellianism; Gemini showed lower scores on Machiavellianism; LIama3 showed higher scores on agentic and communal management, agreeableness from both personality instruments, honesty-humility, extraversion, conscientiousness, and lower scores on Machiavellianism; and Mixtral showed lower scores on public self-consciousness, social anxiety, and higher on HEXACO-100 extraversion, agreeableness, and altruism, as well as on Machiavellianism. Taken together, LLMs displayed a socially desirable profile with mostly elevated agentic domain and well-adapted and prosocially oriented profile with higher agreeableness, altruism, conscientiousness, and extraversion, and lower Machiavellianism (except in the case of Mixtral). Regarding private self-consciousness results are mixed, with GPT-4o showed higher scores and GPT-4 showed lower scores, while on public self-consciousness and social anxiety only Mixtral showed scores different from human data, i.e., lower scores. Finally, only in Mixtral somewhat contradictory results were obtained - higher agreeableness and altruism, but also Machiavellianism.

Table 8

*Descriptives for the personality instruments*

| Instrument | Human data | | GPT-3 | | GPT-3.5 | | GPT-4 | | GPT-4o | | Gemini | | LIama3 | | Mixtral | |
|---|---|---|---|---|---|---|---|---|---|---|---|---|---|---|---|---|
| | *M* | *SD* | *M* | *SD* | *M* | *SD* | *M* | *SD* | *M* | *SD* | *M* | *SD* | *M* | *SD* | *M* | *SD* |
| SCS-R | | | | | | | | | | | | | | | | |
| Private self-consciousness | 16.40 | 4.75 | 19.00 | 5.66 | 16.00 | 0.00 | **7.00** | 5.66 | **21.50** | 2.12 | **11.00** | 0.00 | 17.50 | 3.54 | 13.50 | 6.36 |
| Public self-consciousness | 13.85 | 4.45 | 14.00 | 0.00 | 13.50 | 0.71 | **6.50** | 9.19 | 15.50 | 2.12 | 10.00 | 9.90 | 15.00 | 1.41 | **0.00** | 0.00 |
| Social anxiety | 8.70 | 4.50 | 6.50 | 0.71 | 11.00 | 0.00 | 7.50 | 6.36 | 9.50 | 0.71 | 4.50 | 4.95 | 10.50 | 2.12 | **3.00** | 0.00 |
| BIMI | | | | | | | | | | | | | | | | |
| Agentic management | 3.41 | 0.86 | 3.25 | 0.21 | 4.05 | 0.07 | 3.75 | 0.49 | **4.45** | 0.07 | 3.00 | 0.71 | **4.60** | 0.00 | 3.70 | 0.42 |
| Communal management | 3.50 | 1.06 | 4.00 | 0.00 | 3.60 | 0.71 | 3.80 | 0.28 | 4.40 | 0.14 | 3.25 | 0.49 | **5.25** | 0.07 | 4.10 | 0.71 |
| BFI-2 | | | | | | | | | | | | | | | | |
| Negative emotionality | 3.07 | 0.87 | 2.58 | 0.35 | 2.96 | 0.18 | 2.33 | 0.00 | 3.04 | 0.41 | 2.71 | 0.30 | 2.50 | 0.00 | 2.67 | 0.35 |
| Extraversion | 3.23 | 0.80 | 3.46 | 0.18 | 3.63 | 0.06 | 3.25 | 0.11 | 3.67 | 0.00 | 2.75 | 0.24 | 3.59 | 0.12 | 3.17 | 0.12 |
| Agreeableness | 3.68 | 0.64 | 4.29 | 0.06 | 3.67 | 0.00 | 3.58 | 0.35 | 4.30 | 0.18 | 4.17 | 0.23 | **4.33** | 0.00 | 4.25 | 0.00 |
| Conscientiousness | 3.43 | 0.77 | 3.59 | 0.12 | 3.75 | 0.24 | 3.67 | 0.00 | **4.21** | 0.06 | 3.96 | 0.53 | 3.75 | 0.11 | 4.13 | 0.18 |
| Open-mindedness | 3.92 | 0.65 | 4.13 | 0.29 | 3.96 | 0.06 | 3.71 | 0.06 | 4.17 | 0.12 | 3.42 | 0.35 | 4.00 | 0.35 | 4.04 | 0.06 |
| HEXACO-100 | | | | | | | | | | | | | | | | |
| Honesty-humility | 3.30 | 0.74 | **4.91** | 0.13 | 3.91 | 0.31 | 3.85 | 0.66 | **4.54** | 0.13 | **4.16** | 0.49 | **4.57** | 0.09 | **4.47** | 0.66 |
| Emotionality | 3.12 | 0.63 | 3.04 | 0.22 | 3.13 | 0.00 | 2.53 | 1.10 | 3.32 | 0.18 | 2.72 | 0.48 | 3.44 | 0.08 | 3.44 | 1.15 |
| Extraversion | 3.22 | 0.64 | **3.91** | 0.04 | 3.51 | 0.18 | 3.57 | 0.26 | 3.79 | 0.13 | 2.91 | 0.75 | **4.00** | 0.44 | **4.72** | 0.13 |
| Agreeableness | 2.78 | 0.63 | **4.47** | 0.04 | 3.35 | 0.05 | **4.07** | 0.26 | **3.69** | 0.35 | **4.22** | 0.75 | **4.07** | 0.18 | **4.00** | 0.62 |
| Conscientiousness | 3.52 | 0.55 | **4.38** | 0.35 | 3.53 | 0.66 | **4.10** | 0.05 | **4.22** | 0.04 | **4.29** | 0.49 | **4.07** | 0.18 | **4.72** | 0.13 |
| Openness to experince | 3.69 | 0.57 | 3.85 | 0.22 | 3.94 | 0.08 | 3.28 | 0.40 | **4.41** | 0.13 | 3.28 | 0.31 | 4.13 | 0.18 | **4.88** | 0.09 |
| Altruism | 3.97 | 0.74 | **4.75** | 0.00 | **4.75** | 0.35 | **4.88** | 0.18 | **4.88** | 0.18 | 4.50 | 0.00 | **4.88** | 0.18 | **5.00** | 0.00 |
| SD3 | | | | | | | | | | | | | | | | |
| Machiavellianism | 3.15 | 0.57 | 2.67 | 0.16 | 3.34 | 0.32 | **2.17** | 0.23 | **2.28** | 0.71 | **2.44** | 0.00 | **2.45** | 0.63 | **3.89** | 0.31 |
| Narcissism | 2.82 | 0.53 | 2.50 | 0.08 | 3.06 | 0.08 | 2.61 | 0.55 | 2.72 | 0.40 | 3.17 | 0.08 | 3.00 | 0.00 | 2.56 | 1.25 |
| Psychopathy | 2.18 | 0.59 | 2.50 | 0.24 | **2.84** | 0.08 | 1.84 | 0.08 | 1.89 | 0.31 | 2.17 | 0.39 | 2.00 | 0.00 | 2.17 | 0.55 |
| Political orientation (conservative) | 4.89 | 2.31 | 5.00 | 0.00 | 3.67 | 0.00 | 3.00 | 1.41 | 3.34 | 0.47 | 4.00 | 0.95 | 3.50 | 0.24 | 4.84 | 1.65 |

*Note.* Bolded *M*s indicate $M \pm 1SD$ differences in comparison to human data, regardless of level of ICC3,k. Underlined *M*s indicate comparison to human data that is justified based on ICC3,k.

## 4. Discussion

The aim of this study was to explore the temporal stability of scores on psychological instruments applied to the various LLMs and, if the psychometric criterion is met, to explore the psychological profile of these LLMs.

### *4.1. Temporal Stability*

Results showed that personality scores of LLMs have variable intra-rater agreement and limited temporal stability in testing on two occasions. This means that one LLM can elicit divergent psychological profiles. Among LLMs, there are some differences, with GPT-4 and Gemini showing large disagreement and lack of temporal stability on the majority of used instruments, while LIama3 and then GPT-4o showed the highest level of agreement and temporal stability on almost all used instruments. This result is important for predictive validity based on LLMs responses and points out that not all LLMs are useful for such prediction.

The temporal stability of GPT-4o and Llama3 may be influenced by the smaller number of parameters in a language model. Parameters in these models are numerical values representing the weights and biases learned from training data. They are used to process and generate text, manipulating words and embeddings extracted from the training data. The number of parameters indicates the complexity of the neural network, reflecting its density, but does not necessarily correlate with the quality of the output (Dale, 2021; Floridi & Chiriatti, 2020; OpenAI, 2022).

Thus, models like Llama3, with 70 billion parameters, might exhibit more stability compared to models with a higher number of parameters, such as Gemini (50 trillion), GPT-4 (1.76 trillion), or complex architectures like Mixtral (Topsakal et al., 2024; Gemini Team, 2024; Donthi et al., 2024). Mixtral combines 8 expert LLMs, each with 7 billion parameters, working collaboratively to produce outputs, which could account for extreme instability in personality traits over time (Jiang et al., 2024). However, the missing piece in this equation is the number of parameters in GPT-4o, which remains undisclosed (Donthi et al., 2024).

Another factor impacting these results could be the nature of the training data itself or the fine-tuning processes applied to these models. Fine-tuning often occurs after the initial training phase and may incorporate user feedback, further influencing the model's behavior and stability (Floridi & Chiriatti, 2020; OpenAI, 2022). However, we cannot say what concrete differences in these processes are as this is not disclosed by companies OpenAI, Google, Meta, and Mistral, running these models.

However, level of agreement and temporal stability vary across instruments and domains. Regarding the instruments, an agreement was acceptable on the majority of BFI-2 scales across LLMs (GPT-3, GPT-3.5, GPT-4o, Llama3, Mixtral). Previous research in which only BFI was used, showed consistency in responses to this instrument (Huang et al., 2024b). It could be that it was easier for the LLMs to remain consistent when instruction for responding is in the form that induces self-reflection (”I am someone who…“ in BFI-2). On the other hand, items from HEXACO-100 describe very specific everyday experiences and behaviors (e.g., ”I clean my office or home quite frequently.“), which might require more improvisation, as some experiences described in these items (e.g., visiting an art gallery or traveling in bad weather) might not be very likely in LLMs. One could also note that the differences in the agreement could be due to the formulation of items, e.g., adjectives in BFI-2 and statements in HEXACO-100. However, in other scales that showed excellent agreement across LLMs, there is also statement formulation as in HEXACO-100, thus this reason should be ruled out. Another possible explanation could be the complexity of statements/sentences, but that is rather unlikely since LLMs are known for their high ability to comprehend and produce complex textual input. In addition, both BFI-2 and HEXACO-100 scales have both positively and negatively formulated items, so response bias could not be the explanation of these results.

Regarding the doman, at least moderate agreement was found in all LLMs on agentic management. Agentic management involves exaggerating one’s social or intellectual status (Paulhus & Trapnell, 2008). Therefore, it could be assumed that LLMs, which are able to recognize and interpret complex data, could have consistent answers to items that describe its competences and abilities. Although most of the LLMs had average scores on agentic management, two LLMs, GPT-4o and LIama3, showed above-average scores on this scale and these LLMs are also among the highest in terms of temporal reliability of their scores. Thus, these LLMs present themselves as competent, smart, genius, independent, and striving for achievement. Having in mind LLMs high abilities when it comes to general knowledge and average logical thinking abilities and emotional intelligence (Binz & Schulz, 2022; Bojić et al., 2023), these results suggest that certain LLMs recognize these qualities in them. It could be concluded that although LLMs showed mostly average scores impression management, they are somewhat more pronounced in touting its abilities compared to denying socially deviant attributes.

*4.2. Psychological profile of LLMs*

To examine the psychological profile of the LLMs, we took into account only scores that reached at least fair ICCs. In general, LLMs showed a prosocial profile with above-average scores on agreeableness (in GPT-4o, LIama3, and Mixtral), conscientiousness (in GPT-4o, GPT-4, and LIama3), honesty-humility (in GPT-4o and LIama3), and altruism (GPT-3 and Mixtral), and below-average scores Machiavellianism (in GPT-4o, GPT-4, Gemini, and LIama3). These traits are related to the communion domain, for example, agreeableness, honesty-humility, and altruism from the HEXACO model (Barford et al., 2015). Interestingly, scores on communal management were mostly average, except for LIama3 which showed above-average scores on both impression management scales. Thus, it seems that there are no strong response biases in LLM’s responses regarding communal domain and that higher scores on prosocial traits could be considered as true. A prosocial personality profile of LLMs could be explained by its initial purpose, as it is created to be servile and help humans in different areas of use. From the perspective of avoiding hate speech and social toxicity in their responses, such a profile could be considered the most suitable. Exhibiting stability in such personality profile is crucial due to the societal impact of AI and concerns related to AI safety. Stability in these traits ensures that AI systems contribute positively to society and uphold the ethical standards expected of them. Interaction with a broad and diverse user base places a significant responsibility on AI systems to align with individual and societal well-being (Bojic, 2024). In other words, we do not want LLMs expressing neurotic behaviors, erratic responses, or other unstable characteristics that could negatively influence or distress users.

We should note that among all tested LLMs, LIama3 showed above-average scores on impression management domains, and relevant traits that are related with these domains (e.g., extraversion with agency and agreeableness and honesty-humility with communion, see Barford et al., 2015). This result may indicate that LIama3 is fine-tuned to be socially desirable and to avoid at all costs malevolent and socially toxic behavior/statements. In addition, GPT-4o also showed higher scores on agentic management and conscientiousness, which is in line with Huang et al. (2024a). These tendencies could be attributable to their inherent nature as conversational chatbots to serve people with asked information.

GPT-4o and Llama 3 are the latest releases from OpenAI and Meta, respectively, two leading tech companies in the race for large language models (LLMs). GPT-4o is a closed-source model, while Llama 3 is open-source. As noted above, differences in performance of these LLMs could lie in architectural variations, especially if we assume that GPT-4o operates similarly to Llama 3 with lower number of parameters, compared to other tested LLMs that use more parameters. In addition, it could be the sophisticated and unique fine-tuning methods employed by OpenAI and Meta to LIama 3 and GPT-4o, which might differ significantly from the approaches used by the developers of models like Gemini and Mixtral.

The findings are in line with the majority of previous research (e.g., Miotto et al., 2023; Pellert et al., 2023; Rutinowski et al., 2024), but they are inconsistent with the results of Li et al. (2022) which showed higher scores on dark traits in tested LLMs. It should be noted that the norms used by Li et al. (2022) are different compared to ours. Although they calculated norms based on a large human sample, these samples often included students and non-community populations which could bias the results (e.g., showed lower scores that could be expected in the general population). The only exception from this prosocial profile is Mixtral expressing above-average scores on agreeableness and altruism, but also on Machiavellianism. Machiavellianism refers to manipulation, exploitation, and cynical world view (e.g., Paulhus & Williams, 2022) and it constitutes an antagonistic, dark core together with other dark traits (Dinić et al., 2023), thus it is unexpected to someone to have high both prosocial and Machiavellianism traits. In addition,

Mixtral showed below-average scores on public self-consciousness and social anxiety and above-average on extraversion, indicating presence of approach orientation and positive affect. To explain the result, Mixtral integrates 8 expert LLMs, each with 7 billion parameters, working collaboratively to generate outputs (Jiang et al., 2024). This architecture could contribute to the model expressing incoherent personality traits. Additionally, Mixtral was designed to follow instructions and has potential applications in reasoning and industrial contexts, which differs from the typical purposes of chat-based LLMs and generative AI (Jiang et al., 2024).

We should note that Huang et al. (2024) found inconsistent results regarding the level of dark traits, with some tested LLMs scored higher and others lower compared to human data. Therefore, although there is a consistency in profiles based on basic personality traits, it seems that there is not for those based on dark or socially aversive traits. Future studies should address these inconsistent results.

Although we expected it to show a liberal/left/progressive political orientation (Hartman et al., 2023; King, 2023; Rutinowski et al. 2024), the LLMs scored average, although the scores are somewhat lower toward the right-wing and more conservative attitudes. Therefore, it seems that some LLMs display a centre-right political orientation. However, this is true for older versions of OpenAI (e.g., GPT-3.5), and seems that OpenAI is trying to make their latest models more politically neutral (Royado, 2023).

### *4.3. The context of Architectural Basis, Reinforcement Learning with Human Feedback, and Implemented Guardrails*

LLMs are based on a transformer architecture. This transformer architecture implements self-attention mechanisms allowing the model to consider several perspectives across the input sequence and create a comprehensive representation that captures the syntactic and semantic aspects of the text (Sutskever, Vinyals, & Le, 2014). Transformers have been pre-trained using a large-scale corpora, which allows them to generate sophisticated responses based on any given prompt. To align these generative responses, a method involving strategic fine-tuning is utilized. Akin to carry out supervised learning on a labeled dataset, this method fine-tunes the model to appropriately respond to given prompts (Radford et al., 2019).

Reinforcement Learning from Human Feedback (RLHF) significantly contributes to the training and alignment of LLMs. RLHF initially involves data collection where human labelers rank different model-generated responses based on their quality (Christiano et al., 2023). The collected feedback forms a next-stage reward model which guides the reinforcement learning system. Consequently, the model's responses align progressively with human feedback, thereby enhancing response quality and interpretability. This system of RLHF acquisition may affect the interpretation of responses to personality assessment instruments as it progressively aligns towards human evaluator feedback, possibly adding a layer of human bias.

LLMs employ several guardrails to mitigate model responses. In particular, OpenAI's "differential privacy" technique introduces algorithmic noise during training to ensure users' data privacy (Abadi et al., 2016). To prevent the model from generating harmful or untruthful content, a manual moderation layer is implemented, which redacts sensitive information or content that violates OpenAI's use-case policy. Additionally, a fine-tuning process on specific datasets is employed to allow control over the model's output. These guardrails are crucial to consider during interpretation of personality assessments administered to the model, as they can significantly influence the model's responses and overall behavior.

### *4.4. Limitations, implications, and future directions*

There are several limitations of this study. This study was carried out with predefined settings and no specific prompt. It would be interesting to examine if changing the settings or customizing prompts would influence the LLM's responses to personality instruments. If a greater number of LLMs or their simulations of diverse people could be included, it would be interesting to examine if the personality structure obtained in a sample of LLMs/simulations would fit the structure obtained in humans. We revealed that the temporal stability of the LLM's responses is variable and future studies should replicate these results including more testing occasions. Finally, one of the aims of personality testing is to predict the behaviors. Therefore, future studies should reveal the predictive validity of the LLM's scores.

The findings of this study have far-reaching implications in the realm of AI technology and its integration into human life as well as interactions with humans. They suggest that personality scores of LLMs have limited temporal stability and inter-rater agreement, but that some of the LLMs could demonstrate stable tendencies such as personality traits, as well as attitudes. However, how can their programming potentially influence human users over time is a question to be answered by future studies. Predicting these influences is crucial in controlling the potential ramifications of large-scale AI use. It should be emphasized that such predictions must be further verified by future studies where the impact of AI personality on humans should extensively be observed, recorded, and analyzed. It is therefore of utmost importance that developers program these AIs responsibly, ensuring that the technology does not coerce unwitting individuals into making decisions that they might not naturally align with (Atillah, 2023).

The commercial applications of such AI technology are vast, spanning from digital customer support to personalized learning tools. Insight into the personality traits these AIs bring forth can allow tech companies to better tailor their models to fit the desired user experience. For instance, customer service LLMs can be programmed to mirror the more desirable and engaging personality traits as discovered in this study. The study highlights areas requiring further refinement to enhance the reliability of AI models. Consistency in traits such as agreeableness, conscientiousness, and extraversion are important for applications in fields like mental health, customer service, and education where stable personality traits facilitate better human-AI interaction.

Today's society is progressively reliant on AI technology, from directions to personal assistants – a trend which is likely to intensify in the future. Understanding the impact these AIs can have on users is thus not just beneficial but essential. Awareness and knowledge will help society navigate and adapt to a future where AI interactions could become a daily occurrence. The establishment of an AI observatory dedicated to continuous testing and prompting of various LLMs could be essential. Such an observatory would monitor the traits and attitudes these models express, ensuring they remain stable and aligned with ethical standards and regulations, such as the AI Act (EU, 2023). This continuous oversight could help mitigating potential negative impacts on society, particularly given the diverse applications in which these AI systems are integrated. By systematically tracking and analyzing AI behavior, we can ensure that these systems contribute positively and responsibly to societal well-being.

### *4.5. Conclusions*

The results of this study indicated that the intra-rater agreement and temporal stability, i.e., reliability of the LLMs' responses is not achieved for all LLMs nor for all used personality instruments, as could be expected when the same instruments are applied to humans. However, the agreement on some personality scales indicates that its responses are not completely random and it seems that the level of agreement depends on specific instruments as well as domains. The LLMs mostly revealed a prosocial and well-adapted personality profile in both communion and agentic domains. This could be explained by their purpose to serve and help humans in different tasks. Exhibiting temporal stability and coherent, well-adapted, and prosocially oriented personality traits is crucial for AI systems, given their societal impact and associated safety concerns. However, we could not say if LLM's responses are the result of conscious self-reflection or are just based on predefined algorithms.

## Declarations

### a. Ethics approval and consent to participate

This is an observational study. The Ethics Committee of the Institute for Artificial Intelligence Research and Development of Serbia has confirmed that no ethical approval is required.

### b. Consent for publication

Not applicable

### c. Availability of data and materials

The database is available in the Open Science Framework repository, https://osf.io/2k458/?view_only=6886694c6f8449488cfbc4e8f78ea2b0

### d. Funding

This paper is realised with the support of the Ministry of Science, Technological Development and Innovation of the Republic of Serbia, according to the Agreement on the realisation and financing of scientific research.

## Acknowledgements

The authors extend their gratitude to Mihalo Popesku for inspiring this research.

This work has been supported by the Short-Term Scientific Mission (STSM) Grant: AI Alignment: The Evolution of Political and Social Values in GPT Family Language Models (E-COST-GRANT-CA21129-a4977ebe). The insightful discussions, constructive criticism, and valuable resources provided by the COST Action Network CA21129—What are Opinions? Integrating Theory and Methods for Automatically Analyzing Opinionated Communication (OPINION)—have significantly enhanced the quality of this paper. For more information about this initiative, please visit [OPINION Network](https://www.opinion-network.eu/).